%% file: arXiv.tex
\documentclass[10pt,twocolumn,letterpaper]{article}

\usepackage[pagenumbers]{wacv}      % To produce the REVIEW version for the algorithms track
\usepackage{graphicx}
\usepackage{amsmath}
\usepackage{amssymb}
\usepackage{booktabs}

\usepackage[pagebackref,breaklinks,colorlinks]{hyperref}

\usepackage[capitalize]{cleveref}
\crefname{section}{Sec.}{Secs.}
\Crefname{section}{Section}{Sections}
\Crefname{table}{Table}{Tables}
\crefname{table}{Tab.}{Tabs.}
\input{custom}
\def\wacvPaperID{493} % *** Enter the WACV Paper ID here
\def\confName{WACV}
\def\confYear{2025}

\begin{document}

\title{Revisiting Semi-supervised Adversarial Robustness \\ via Noise-aware Online Robust Distillation}

\author{
Tsung-Han Wu$^1$\thanks{Work done while at National Taiwan University (NTU).} \qquad Hung-Ting Su$^2$ \qquad Shang-Tse Chen$^2$ \qquad Winston H. Hsu$^{2,3}$
\\
\\
$^1$University of California, Berkeley \quad $^2$National Taiwan University \quad $^3$Mobile Drive Technology
}
\maketitle

\begin{abstract}
\input{sections/0_abstract}
\end{abstract}

\section{Introduction}
\input{sections/1_introduction}

\section {Related Work}

\input{sections/2_relatedwork}

\section {Method}
\input{sections/3_method}

\section {Experiments}

\input{sections/4_experiment}

\section {Conclusion}
\input{sections/5_conclusion}

\clearpage

\section*{Acknowledgement}
This work was supported in part by the National Science and Technology Council under Grants MOST 112-2634-F-002-006, MOST 110-2222-E-002-014-MY3, NSTC-112-2634-F-002-002-MBK, and by Center of Data Intelligence: Technologies, Applications, and Systems, National Taiwan University under Grant NTU-113L900903.

%%%%%%%%% REFERENCES
{\small
\bibliographystyle{ieee_fullname}
% \bibliography{egbib}
\bibliography{ref.bib}
}
\input{sections/X_supp_arXiv}

\end{document}

%% file: custom.tex
\usepackage{multirow}
\usepackage{xcolor}
\usepackage{booktabs}
\usepackage{setspace}
\usepackage{amsfonts}
\usepackage{amsmath}
\usepackage{colortbl}
\usepackage{wrapfig}

\definecolor{gray}{rgb}{0.5,0.5,0.5} 
\definecolor{frenchblue}{rgb}{0.0, 0.45, 0.73}
\definecolor{gray}{rgb}{0.5,0.5,0.5} 
\definecolor{green}{rgb}{0, 0.4, 0} 
\definecolor{orange}{rgb}{1, 0.5, 0} 	
\definecolor{mahogany}{rgb}{0.75, 0.25, 0.0}
\definecolor{purple}{rgb}{0.6, 0, 0.6}
\definecolor{darkgreen}{rgb}{0, 0.4, 0.4} 
\definecolor{red}{rgb}{1.0, 0, 0}
\definecolor{plotpurple}{rgb}{0.2353, 0.2, 0.90196}
\definecolor{plotorange}{rgb}{1.0, 0.6, 0.2}
\definecolor{plotgreen}{rgb}{0.2, 0.784313, 0.2}
\definecolor{plotred}{rgb}{1.0, 0.2, 0.392}
\definecolor{LightCyan}{rgb}{0.88,1,1}

\newcommand{\tabincell}[2]{\begin{tabular}{@{}#1@{}}#2\end{tabular}}

\let\titleold\title
\renewcommand{\title}[1]{\titleold{#1}\newcommand{\thetitle}{#1}}
\def\maketitlesupplementary
   {
   \newpage
       \twocolumn[
        \centering
        \Large
        \textbf{\thetitle}\\
        \vspace{0.5em}Supplementary Material \\
        \vspace{1.0em}
       ] %< twocolumn
   }

%% file: sections/0_abstract.tex
The robust self-training (RST) framework has emerged as a prominent approach for semi-supervised adversarial training. To explore the possibility of tackling more complicated tasks with even lower labeling budgets, unlike prior approaches that rely on robust pretrained models, we present SNORD -- a simple yet effective framework that introduces contemporary semi-supervised learning techniques into the realm of adversarial training. By enhancing pseudo labels and managing noisy training data more effectively, SNORD showcases impressive, state-of-the-art performance across diverse datasets and labeling budgets, all without the need for pretrained models. Compared to full adversarial supervision, SNORD achieves a 90\% relative robust accuracy under $\ell_{\infty} = 8/255$ AutoAttack, requiring less than 0.1\%, 2\%, and 10\% labels for CIFAR-10, CIFAR-100, and TinyImageNet-200, respectively. Additional experiments confirm the efficacy of each component and demonstrate the adaptability of integrating SNORD with existing adversarial pretraining strategies to further bolster robustness.

%% file: sections/1_introduction.tex
The growing usage of machine learning models in safety-critical areas, like facial recognition systems, has highlighted their vulnerability to adversarial attacks \cite{rozsa2019facial}. Despite advancements in adversarial defense strategies \cite{goodfellow2014explaining, madry2018towards, zhang2019theoretically, zhang2021_GAIRAT, croce2020reliable}, reaching robustness remains challenging under limited labeled data \cite{schmidt2018adversarially}. This issue has sparked a shift in research focus toward the semi-supervised learning (SSL) paradigm to enhance model robustness.

\begin{figure}[!t]
    \centering
    \includegraphics[width=\linewidth]{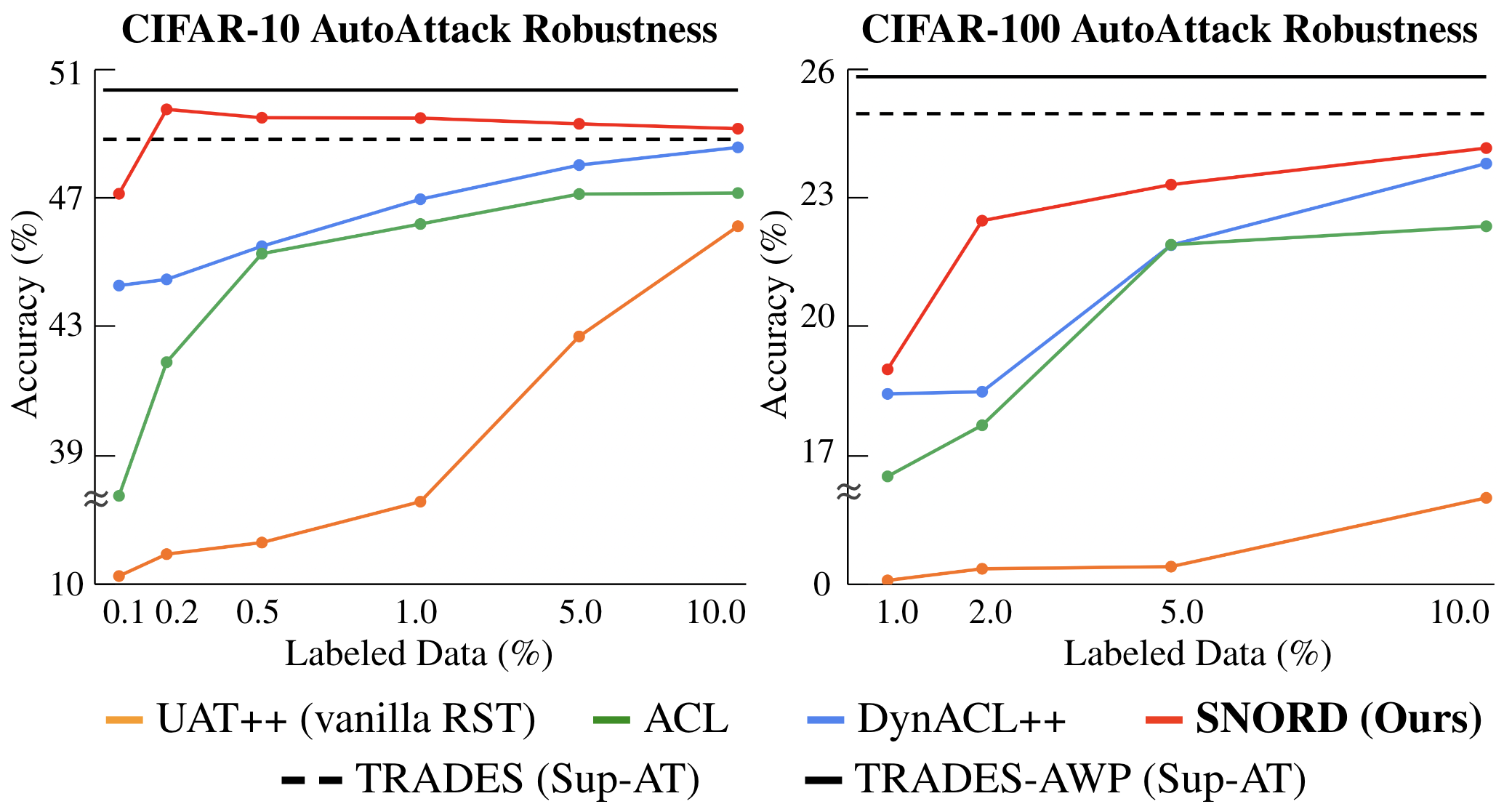}
    \caption{\textbf{Performance comparison of SSL adversarial training techniques.} The RST pipeline is widely adopted by current SSL adversarial training methods. While ACL and DynACL++ incorporate robust pretrained models to enhance the basic RST, their achievements remain suboptimal due to the intrinsic limitations of the RST (further elaborated in \Cref{fig:fig15}). After addressing these issues, our SNORD framework outperforms all these methods by a large margin across diverse labeling budgets and datasets. Notably, on the CIFAR-10 dataset, SNORD attains comparable results to fully adversarial training methods like TRADES and TRADES-AWP but requires \textbf{only 0.2\% of the labeling effort}.}
    \label{fig:fig1}
\end{figure}

\begin{figure*}[!t]
    \centering
    \includegraphics[width=\linewidth]{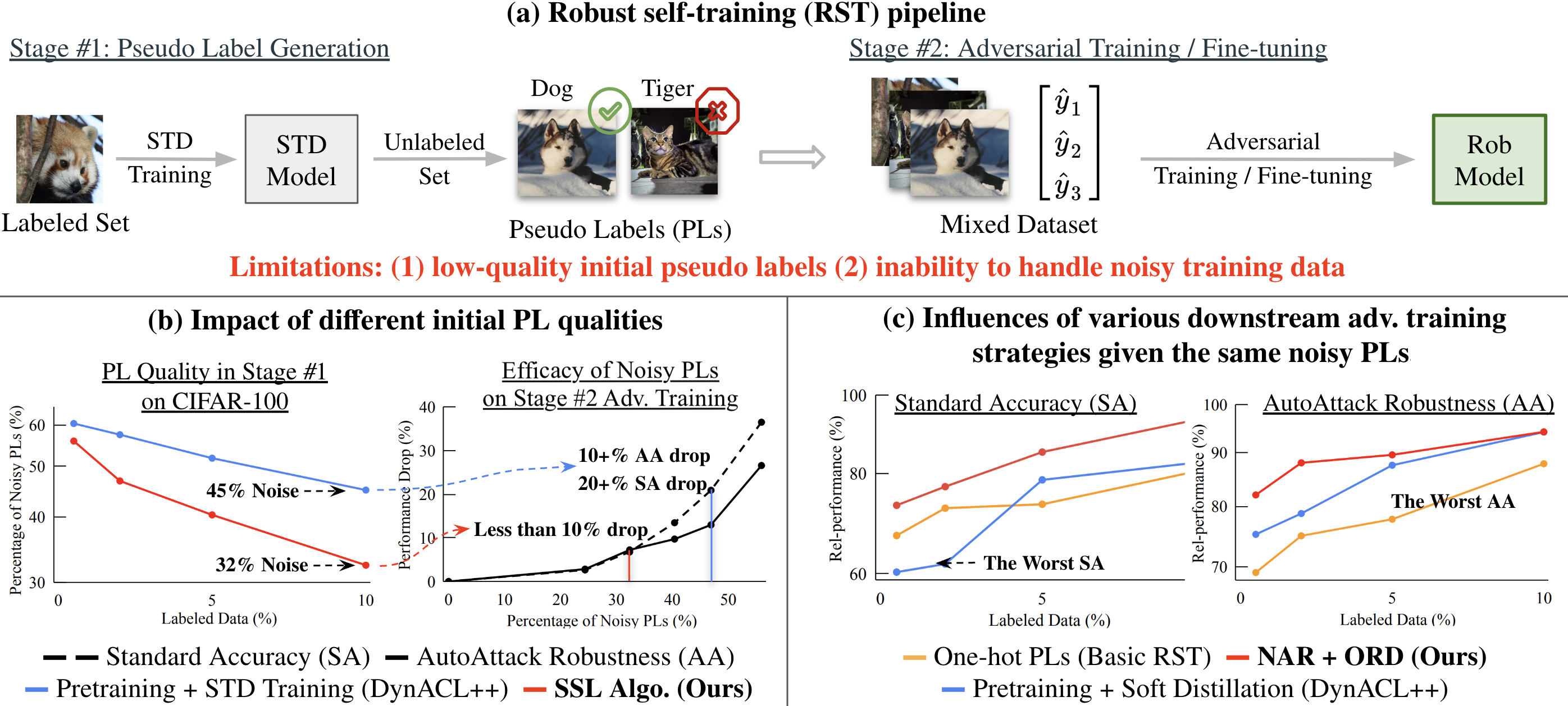}
    % \vspace{-1.5em}
    \caption{\textbf{Revisiting SSL Adversarial Training.} (a) Addressing the limitations of the two-stage RST method involves enhancing initial pseudo label (PL) quality and efficiently managing noisy data in downstream adversarial training. (b) On CIFAR-100, the state-of-the-art RST-based method (DynACL++) still generates lower-quality PLs initially, leading to suboptimal Standard Accuracy (SA) and Adversarial Accuracy (AA) after conventional adversarial training, compared to a fully adversarially trained oracle model. Our approach, utilizing an advanced SSL algorithm for PL generation, significantly improves performance under identical downstream training conditions. (c) We analyze the impact of different downstream adversarial training strategies on CIFAR-100 with equivalent noisy PLs. The y-axis indicates the relative performance compared to the oracle case—the performance under adversarial training using fully labeled data. The results show that the basic RST method with hard labels underperforms due to inaccurate PLs, resulting in the worst AA. Strategies combining adversarial pretraining with soft distillation maintain higher AA but at the cost of reduced SA in a low labeling regime. Conversely, our proposed noise-aware rectification and online robust distillation effectively overcome these issues, achieving superior SA and AA. Further details about our method are provided in \Cref{sec:method}.}
    
    \label{fig:fig15}
\end{figure*}

In the realm of SSL adversarial training, the robust self-training (RST) pipeline has become increasingly popular \cite{carmon2019unlabeled, alayrac2019labels, zhai2019adversarially}. As shown in \Cref{fig:fig15} (a), this two-stage pipeline combines pseudo label generation via standard-trained models and traditional adversarial training methods. While several research has focused on improving RST through the integration of adversarial pretraining methods recently \cite{ACL2020, luo2023rethinking}, \Cref{fig:fig1} reveals that these enhancements still fall short in terms of performance, particularly in scenarios under low labeling regime.

In this work, we revisit these RST-based approaches and identify two critical but frequently overlooked issues: the production of low-quality pseudo labels and the difficulty in managing noisy training data. To begin with, our analysis in \Cref{fig:fig15} (b) indicates that an excess of noisy pseudo labels (over 40\%) in complex datasets like CIFAR-100 can lead to a significant decrease in model robustness and accuracy. Yet, with less than 10\% labels, current RST-based methods often generate pseudo labels with error rates exceeding 45\% in the initial stages, which is alarmingly high.

Additionally, we observe that existing methods struggle to learn from noisy training samples during the subsequent adversarial training phase, as illustrated in \Cref{fig:fig15} (c). Specifically, the basic RST method (UAT++) \cite{carmon2019unlabeled} exhibits the lowest robustness, attributed to training with inaccurate one-hot pseudo labels. While there have been efforts to employ adversarial pretrained weights as initialization and soft distillation loss on unlabeled data \cite{ACL2020, luo2023rethinking}, despite exhibiting reasonable robustness, these methods have proven the worst standard accuracy with less than 5\% labeled data.

To address these issues, we introduce a novel framework for SSL adversarial training, dubbed SNORD (\textbf{S}emi-supervised \textbf{N}oise-aware \textbf{O}nline \textbf{R}obust \textbf{D}istillation). SNORD is inspired by recent successes in standard SSL. Firstly, we utilize an off-the-shelf SSL algorithm \cite{sohn2020fixmatch} for pseudo label generation, significantly improving label quality, as shown \Cref{fig:fig15} (b). Secondly, we present a novel noise-aware rectification strategy and an online robust distillation mechanism for the subsequent adversarial training process. The former further enhances pseudo label quality for downstream training by integrating entropy minimization techniques \cite{lee2013pseudo, miyato2018virtual} and accounting for noise in both labeled and unlabeled data. The latter, inspired by consistency regularization, allows the downstream adversarially trained model to learn from labels across different epochs. The whole pipeline of our SNORD is depicted in \Cref{fig:fig2}.

Through extensive experiments on CIFAR-10/100 datasets \cite{krizhevsky2009learning} and the TinyImageNet-200 dataset \cite{le2015tiny}, we showcase the superior performance of SNORD with a large margin compared to prior practices across varying labeling budgets. Remarkably, SNORD achieves 90\% of the robust accuracy relative to the oracle performance under an $\ell_{\infty} = 8/255$ AutoAttack, as benchmarked against the theoretically ideal, fully supervised method, TRADES \cite{zhang2019theoretically}. This high level of performance is accomplished with considerably fewer labels—less than 0.1\%, 2\%, and 10\% of the total labels for the CIFAR-10, CIFAR-100, and TinyImageNet-200 datasets, respectively. Further experiments underscore the effectiveness and robustness of each component within SNORD, and demonstrate its compatibility and enhanced performance when combined with existing adversarial pretraining strategies. In summary, our contributions are: 

\begin{enumerate}
    \item We are the first to identify the two critical yet previously overlooked issues related to pseudo labels within the widely adopted RST pipeline.
    \item We present SNORD, a simple, effective, and general SSL adversarial training framework to address the above challenges.
    \item Our SNORD framework not only achieves SOTA performance in various adversarial robustness benchmarks but also demonstrates compatibility with existing adversarial pretraining methods.
\end{enumerate}

%% file: sections/2_relatedwork.tex
\label{sec:relatedwork}

\subsection{Semi-supervised Learning (SSL)}

SSL has become increasingly popular due to its capacity for leveraging abundant unlabeled data to enhance model performance. Within the realm of SSL, existing strategies can be broadly categorized into two primary domains: entropy minimization and consistency regularization.

In the context of entropy minimization, methods often assume that a classifier's decision boundary should steer clear of high-density regions in the data distribution. To enact this principle, \cite{lee2013pseudo} introduced the ``pseudo-labeling" technique, a straightforward yet remarkably effective approach utilizing one-hot encoding. Recent advancements \cite{berthelot2019mixmatch,miyato2018virtual} have further refined this approach through label sharpening, enhancing label distributions via softer labels.

Conversely, consistency regularization leverages data augmentation to reinforce the SSL process, ensuring a classifier's output class distribution remains consistent for unlabeled instances even post-augmentation. Several methods and loss functions \cite{rasmus2015semi, sajjadi2016regularization,Berthelot2020ReMixMatch:, xie2020unsupervised, tarvainen2017mean} have been proposed to realize this idea in diverse ways.

While recent studies have achieved remarkable success by combining these strategies for standard image classification, such as FixMatch \cite{sohn2020fixmatch}, their application to the domain of SSL adversarial training remains limited. In this work, we introduce a pioneering unified framework that integrates both entropy minimization and consistency regularization into the field of adversarial robustness. This framework not only significantly surpasses existing baselines but also offers a novel perspective to the field.

\subsection{Adversarial Robustness}
% Introduction
Research in adversarial robustness can be broadly categorized into two fronts: attacks and defenses. Adversarial attacks aim to craft adversarial samples that are misclassified by models through introducing minimal perturbations to benign data, while defensive approaches seek to enhance model robustness against such attacks. Over the past few years, numerous classical attacks, such as Fast Gradient Sign Method (FGSM) \cite{goodfellow2014explaining} Projected Gradient Descent (PGD) \cite{madry2018towards}, and AutoAttack (AA) \cite{croce2020reliable}, have generated adversarial examples by back-propagating loss functions. On the other hand, defensive methods have employed techniques like obfuscated gradients \cite{athalye2018obfuscated} or various adversarial training strategies \cite{madry2018towards,zhang2019theoretically,zhang2021_GAIRAT,wu2023annealing}.

% Noise in adversarial training
Recent studies have delved into the impact of noisy training data on adversarial training. \cite{chen2021robust, dong2022label, zhu2021understanding} mitigated mismatched distribution noise between benign and adversarial samples by applying knowledge distillation loss to smoothed label distributions. In contrast, \cite{zhang2022noilin} introduced a noise injection mechanism to counter robust overfitting. Unlike these endeavors, which focus on noisy labels within supervised learning contexts, our novel noise-aware label rectification strategy addresses inaccurate pseudo label noise specific to the SSL adversarial training paradigm. This addresses a distinct challenge and contributes to enhanced robustness in scenarios with limited labeled data.

\subsection{Semi-supervised Adversarial Robustness}
\cite{schmidt2018adversarially} demonstrated that increasing the amount of labeled data can bolster the adversarial robustness of models. This insight led to the emergence of research on achieving robustness with limited labeled data, termed semi-supervised adversarial robustness. Early studies \cite{carmon2019unlabeled,alayrac2019labels,zhai2019adversarially,li2022semi} introduced the two-stage Robust Self-Training (RST) pipeline, involving pseudo label generation from a standard-trained model in the initial stage and adversarial training on the entire dataset in the subsequent stage.

Despite the simplicity of RST, it often experiences significant performance degradation when labeled data is scarce (e.g., \textless 10\% on CIFAR-10) \cite{gowal2021selfsupervised}. To address this, several researchers have developed self-supervised adversarial training strategies to obtain pretrained models with robust feature representations \cite{Naseer_2020_CVPR,hendrycks2019using,chen2020self,chen2020adversarial,gowal2021selfsupervised, luo2023rethinking,10.1007/978-3-031-20056-4_42,fan2021when,NEURIPS2020_1f1baa5b,NEURIPS2020_c68c9c82}. These techniques have achieved over 85\% robustness compared to fully supervised methods on relatively simple datasets like CIFAR-10, using only 1\% to 10\% annotations of the entire training set.

Building upon this line of research, we observe that existing RST-based methods struggle with complex tasks or extremely limited labeling scenarios due to the low-quality pseudo labels in the initial stage and the inability to handle noisy training data in the subsequent stages. To address these challenges, we propose a novel approach that seamlessly integrates contemporary SSL techniques into the realm of adversarial training. Our proposed framework not only achieves state-of-the-art across established benchmarks but is also applicable to existing adversarial pretrained models, signifying a significant advancement in the domain of semi-supervised adversarial robustness.

%% file: sections/3_method.tex
\label{sec:method}

\subsection{Overview}

In a semi-supervised adversarial robustness problem, our goal is to create a robust model that can withstand adversarial attacks, given only a small labeled dataset and a larger pool of unlabeled data. Specifically, we focus on countering $\epsilon$-tolerant $L_\infty$ attacks on image classification, where the maximum perturbation $\delta$ added to the input $x$ is constrained by $||\delta||_\infty \leq \epsilon$. Our overall objective is to minimize the following expression:

\begin{equation}
    \min_{f_{rob}} \mathbb{E}_{x \in D} \left[\max_{||\delta||_\infty \leq \epsilon} \mathcal{L}(f_{rob}(x + \delta), y)\right],
\end{equation}
where $f_{rob}$ denotes the parameters of the robust model, $D$ is the dataset, and $\mathcal{L}(f_{rob}(x + \delta), y)$ is a 0/1 error function measuring the model's performance under attack.

\begin{figure*}[!t]
    \centering
    \includegraphics[width=\linewidth]{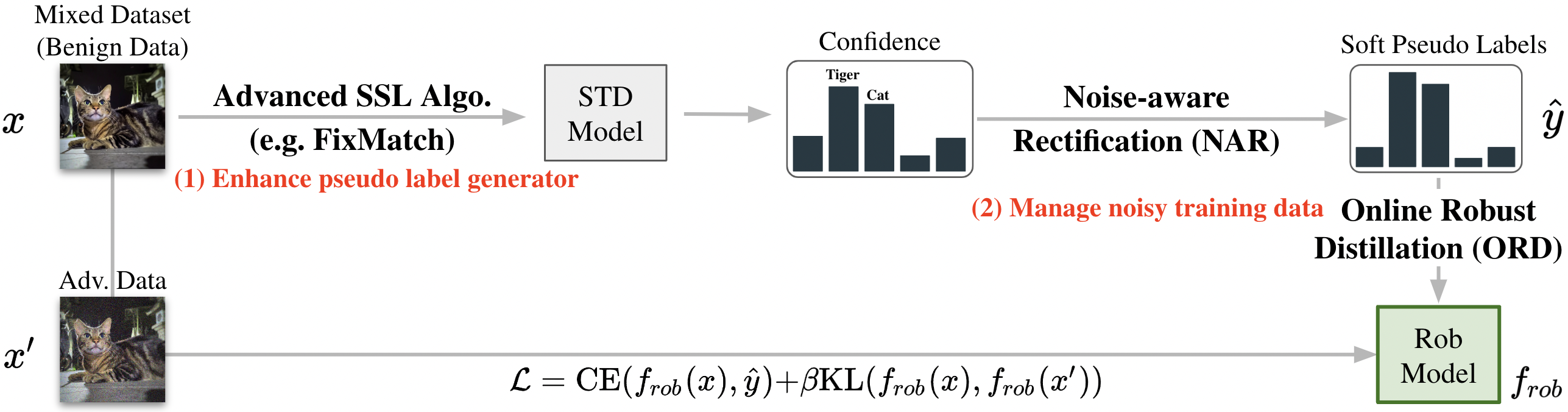}

    \caption{\textbf{Semi-supervised Noise-aware Online Robust Distillation (SNORD).} Our SNORD framework aims to address two important yet previously overlooked issues associated with noisy pseudo labels in the field of semi-supervised adversarial training. Firstly, it employs advanced SSL algorithms to improve the quality of pseudo labels (\Cref{ssec:ssl_algo}). Secondly, we introduce Noise-aware Rectification (\Cref{ssec:nar}) and Online Robust Distillation (\Cref{ssec:ord}) to enhance the learning capabilities of the downstream adversarial robust model in the context of handling noisy estimated pseudo labels.}

    \label{fig:fig2}
\end{figure*}

% Our porposal
In this setting, existing semi-supervised adversarial training methods have some limitations, notably regarding the quality of generated pseudo labels in the initial stage and their ability to handle noisy data in later stages, which often leads to sub-optimal robustness. In response to these challenges, we propose a novel framework named Semi-supervised Noise-aware Online Robust Distillation (SNORD) as shown in \Cref{fig:fig2}, which consists of three key modules: (1) a reliable pseudo label generator, (2) the Noise-aware Rectification (NAR) strategy, and (3) the Online Robust Distillation (ORD) mechanism. We elaborate on each component in the following subsections.

\subsection{Enhancing Pseudo Label Generation}
\label{ssec:ssl_algo}

The cornerstone of a robust self-training process heavily relies on the precision of pseudo labels in its initial phase \cite{gowal2021selfsupervised}. Conventional approaches typically employ supervised training coupled with a pretrained model to derive the pseudo label generator $f_{std}$ \cite{ACL2020}. While these methods are able to achieve 85\% pseudo label accuracy with just 1\% annotations on CIFAR-10, their potential to produce high-fidelity labels in more complex classification tasks or smaller amount of labels remains largely unexplored. Evident from the CIFAR-100 analyses in \Cref{fig:fig15} (b), as the ratio of inaccurate predicted pseudo labels grows from 30\% to around 50\%, the final standard accuracy and adversarial robustness plummet by around 30\% and 20\%, respectively. Notably, given that the widely-adopted model (ResNet-18) consistently demonstrates a classification error surpassing 20\% on CIFAR-100 \cite{zagoruyko2016wide}, this issue can become quite serious during the subsequent adversarial training phase, not to mention when dealing with even scarcer labeling resources or more complex classification task, such as the case of TinyImageNet-200.

To acquire more accurate pseudo labels across varied scenarios, we embrace advanced semi-supervised learning (SSL) algorithms like FixMatch \cite{sohn2020fixmatch} and ReMixMatch \cite{Berthelot2020ReMixMatch:}. After establishing the SSL-trained pseudo label generator $f_{std}$, we feed each benign image $x$ through $f_{std}$ to yield an estimated label distribution noted as $p(x) = \mathrm{softmax}(f_{std}(x))$. These estimations serve as the foundation for training the downstream adversarial robust model $f_{rob}$, which we elaborate on in the subsequent section.

\subsection{Noise-aware Rectification}
\label{ssec:nar}
As shown in \Cref{fig:fig15} (c), on unlabeled data, prior methods either directly leveraged the soft predicted distribution as targets without applying entropy minimization technique or training the robust model with hard labels while overlooking inaccurate guidance. To enhance the use of the estimated label distribution in unlabeled data for subsequent adversarial training, we introduce a novel approach to improve pseudo labels, termed Noise-Aware Rectification (NAR). This method focuses on three key areas: mitigating the effects of mismatched label distribution noise in adversarial training, addressing inherent noise in incorrect predicted pseudo labels, and integrating the label sharpening techniques commonly used in the field of standard semi-supervised learning. The NAR strategy is mathematically represented as follows:
\begin{equation}
\hat{y}=\begin{cases}
\lambda \cdot p(x) + (1-\lambda) \cdot y_{GT}, & \text{if $x \in D_L$}\\
\lambda \cdot p(x) + (1-\lambda) \cdot y_{PL}, & \text{if $x \in D_U$.}\
\end{cases}
\end{equation}
In this formula, $D_L$ and $D_U$ indicate labeled and unlabeled datasets, respectively. $y_{GT}$ represents the one-hot ground truth labels, $\lambda$ is a label sharpening factor that balances the predicted distribution and one-hot labels, and $y_{PL}$ is a one-hot label vector sampled from the predicted probability distribution $p(x)$. The sampling process for each dimension $i$ of $y_{PL}$ in a training batch is defined as:
\begin{equation}
y_{PL}^i=\begin{cases}
1, & \text{if $i$ is sampled from $p(x)$,}\\
0, & \text{Otherwise.}
\end{cases}
\end{equation}

Our design choices stem from three main factors. Firstly, recent studies point out a drawback of using one-hot label assignments for adversarial training, as adversarial perturbations can introduce label noise by distorting data semantics \cite{dong2022label}. To counter this, we implement a label fusion technique that merges one-hot labels $y_{GT}$ with the estimated distribution $p(x)$, resulting in a smoother label distribution.

Secondly, drawing inspiration from the success of entropy minimization in standard SSL tasks \cite{berthelot2019mixmatch}, we apply a similar technique to refine the label distribution $p(x)$ for the unlabeled data. This approach considers the potential inaccuracies of initial pseudo-label generators and opts for sampling a one-hot label $y_{PL}$ from the distribution $p(x)$ instead of using a traditional $\mathrm{argmax}(\cdot)$ function. This strategy not only improves the label distribution for the unlabeled subset but also accounts for noise.

Finally, our approach applies the same label rectification process to both labeled and unlabeled data, using a single shared parameter. This contrasts with previous methods \cite{ACL2020, carmon2019unlabeled} and reduces the need for hyperparameter tuning while ensuring a balanced treatment of both data subsets. A detailed discussion and comparison of our NAR strategy are presented in \Cref{sec:abl}.

\subsection {Online Robust Distillation}
\label{ssec:ord}
While the combination of an improved pseudo label generator and a sophisticated noise-aware rectification strategy lays the foundation for the second-stage adversarial training, this adversarial training method cannot demonstrate the full potential of our approach. To further elevate its robustness, we introduce the concept of consistency regularization, a prevailing SSL strategy, into the realm of adversarial training. Drawing inspiration from the well-established research \cite{tarvainen2017mean, zi2021revisiting}, we would like to leverage the efficacy of techniques such as label smoothing and ensemble methods across different epochs to enhance standard accuracy and adversarial robustness. In this work, we introduce an innovative online distillation mechanism that trains the robust model while concurrently updating the pseudo-label generator. In this process, we compute the estimated probability distribution at epoch $t$:
\begin{equation}
  p^t(x) = f_{std}^{t}(\alpha(x))
\end{equation}
where $\alpha(\cdot)$ denotes a weak augmentation function, strategically introduced to diversify labels and further fortify robustness. To ensure training stability, the labels of the robust student undergo updates following the training of the pseudo label generator for $T$ epochs. This dynamic knowledge distillation process in an online fashion empowers the teacher to offer more diverse and reliable guidance to its robust student, invariably leading to superior outcomes. 

With an enhanced pseudo label generator, advanced label rectification strategy, and online distillation mechanism, the overall loss function for our holistic SNORD framework is as follows:
\begin{equation}
\mathcal{L} = \mathrm{CE}(f_{rob}(x), \hat{y}) + \beta \mathrm{KL}(f_{rob}(x), f_{rob}(x')),
\end{equation}
where $\hat{y}$ is computed with equations 2, 3, and 4. 

%% file: sections/4_experiment.tex
\subsection{Experimental Settings}

We describe the datasets and experimental protocols in this section. Implementation details including hyper-parameter settings are reported in the supplementary material.

\vspace{0.75em}

\noindent {\textbf{Datasets.}} Our experiments were conducted with the widely-used CIFAR-10/100 datasets \cite{krizhevsky2009learning} and the TinyImageNet-200 dataset \cite{le2015tiny}. Following the established protocols from previous SSL adversarial training studies \cite{luo2023rethinking, ACL2020}, for CIFAR-10 and CIFAR-100, we evaluated our models using the official test set, while the official training set was divided into a 9:1 ratio for training and validation. For TinyImageNet-200, we directly used the official training, validation, and test set in the experiment. To create SSL settings, we randomly partitioned the training set into labeled and unlabeled portions to meet our specific experimental requirements following \cite{luo2023rethinking, ACL2020}. To mitigate potential class-imbalanced issues during training, we made sure to distribute the images of each class evenly across the splits. In cases where the number of images couldn't be divided exactly, a small difference of at most one image was allowed.

\vspace{0.75em}

\input{tables/big_tab_C10}

\input{tables/big_tab_C100}

\noindent {\textbf{Training and Evaluation Protocols.}} For fair comparisons, we employed the widely-used ResNet-18 \cite{he2016deep} model architecture for all experiments. During training, all methods were allowed to optimize the model while considering a maximum perturbation of $L_{\infty} = 8/255$. The maximum number of perturbation steps was limited to 10 and the step size of each perturbation is set to $\alpha = 2/255$.

For our evaluation, we adhered to the protocols established in prior works \cite{luo2023rethinking, ACL2020}, utilizing three standard metrics for a thorough assessment: standard accuracy (SA), PGD-20 robust accuracy (RA) \cite{madry2018towards}, and AutoAttack (AA) accuracy \cite{croce2020reliable}. PGD-20 refers to the robustness of the model when challenged with the Projected Gradient Descent method, a widely recognized adversarial attack approach, using 20 iterations. AutoAttack (AA), on the other hand, is an ensemble of diverse attack methods designed to provide a more comprehensive and stringent test of model robustness. In both metrics, the maximum perturbation was bounded by $L_{\infty}=8/255$ across all three datasets: CIFAR-10, CIFAR-100, and TinyImageNet-200. To gain a comprehensive understanding of the problem, we evaluated all SSL adversarial training methods across a range of labeling budgets. Specifically, the labeling budgets for CIFAR-10, CIFAR-100, and TinyImageNet-200 varied from $0.1\% \sim 10 \%$, $1\% \sim 10 \%$, and $10\% \sim 20\%$, respectively.

\subsection{Main Results}
\label{sec:result}

We compared our method with three semi-supervised adversarial training baselines. Among them, UAT++ \cite{alayrac2019labels} is the basic RST method, while ACL \cite{ACL2020} and DynACL++ \cite{luo2023rethinking} involve a self-supervised adversarial pretraining stage followed by a few RST-based finetuning steps. As UAT++ did not release the source code, we re-implemented the method and reported the number using our code. For ACL and DynACL++, we leveraged their provided pretrained model and source code to obtain the result. We reported the maximum value of their official number and our finetuned results.

\vspace{0.75em}

\noindent {\textbf{CIFAR-10.}} The results in \Cref{tab:main_result} underscore the remarkable efficacy of our SNORD framework over existing baselines across a spectrum of labeling budgets. In the context of scarce labeling resources (\textless 1\%), a regime in which conventional methods falter in generating high-quality pseudo labels, SNORD exhibits an impressive advancement of around 3\% in AA when compared to all established baselines. This highlights the advantages inherent in our approach of leveraging sophisticated SSL algorithms to derive pseudo labels, as opposed to relying solely on pretrained models within standard training paradigms.

On the other hand, even when at least 1\% labeled data is available—enough for baseline methods to generate satisfactory pseudo labels—SNORD not only slightly outperforms these baselines in terms of AA, but also provides a substantial improvement in SA. The huge gain actually comes from the benefits of our NAR and ORD modules, which collectively facilitate the effective management of noisy training data stemming from inaccurate pseudo label while also harnessing robust entropy minimization techniques to bolster standard accuracy.

Notably, even with a mere 0.2\% labeled data (equivalent to only 9 labeled images per class), SNORD achieves an incredible 49\% AA robustness. This impressive achievement underscores the superior capabilities of our method in harnessing minimal labeled data to rival the outcomes of extensively supervised methods such as TRADES and TRADES-AWP as showcased in \Cref{fig:fig1}.

\vspace{0.75em}

\noindent {\textbf{CIFAR-100.}} As presented in \Cref{tab:c100_result}, as the CIFAR-100 is much more difficult than CIFAR-10, the performance of the baseline vanilla RST method (UAT++) is notably bad, failing to exceed 20\% AA with even with 10\% available annotated data. In such a complicated task, while the integration of pretrained models indeed provides improvement compared with UAT++, our SNORD framework continues to outperform these approaches across a diverse range of labeling budgets. Echoing the results seen in CIFAR-10, SNORD consistently yields significantly improved standard accuracy (SA) over baseline methods, particularly in scenarios with ample labeling resources.

It is important to note that in scenarios where only 1\% of labeled data is available (equivalent to only 4 to 5 labeled images per class), the prior state-of-the-art DynACL++ method manages to achieve results that closely approximate those of SNORD. This situation can be attributed to the relatively high proportion of inaccurate pseudo labels generated by SNORD, thus obtaining only compromised results compared with the performance under 2\% labels.

\vspace{0.75em}

\noindent {\textbf{TinyImageNet-200.}} As depicted in \Cref{tab:tin_result}, our SNORD method consistently shows exceptional performance, even when faced with a more challenging dataset, as compared directly to the basic RST method (UAT++). It is worth highlighting that the utilization of the supervised TRADES method yields SA and AA scores of only 48.49 and 17.35, respectively. In stark contrast, SNORD achieves results that outperform 95\% of it with merely 20\% labeled data.

\input{tables/big_tab_TIN}

\vspace{0.5em}

\noindent\textbf{Comparison with Full Adversarial Training.} We also compare the proposed SNORD pipeline with the state-of-the-art fully adversarial training method, ADR \cite{wu2023annealing}, which utilizes the entire set of labeled data. As depicted in \Cref{tab:sup_comp}, SNORD, even with 10 times fewer labels, achieves impressive results in both Standard Accuracy (SA) and AutoAttack (AA) across different datasets. Most notably, SNORD substantially narrows the AA performance gap to less than 5\% when compared to the leading full adversarial training method. Furthermore, SNORD significantly surpasses existing semi-supervised adversarial training methods, particularly with the large-scale TinyImageNet-200 dataset and in the low-label regimes of CIFAR-10/100. These findings underscore SNORD's remarkable efficacy and contribution in the realm of SSL adversarial training.

\begin{table}[h]
\setlength\tabcolsep{3pt}
\resizebox{1.0\linewidth}{!}{
\begin{tabular}{ccccccc}
    \toprule
    \multicolumn{1}{c}{\multirow{2}{*}{Method (labels)}} & \multicolumn{2}{c}{CIFAR-10} & \multicolumn{2}{c}{CIFAR-100} & \multicolumn{2}{c}{TinyImageNet} \\
    \multicolumn{1}{c}{} & SA & AA & SA & AA & SA & AA \\
    \midrule
    ADR \small{(100\%)}&82.41&50.38&56.10&26.87&48.19&19.46\\
    \midrule
    Prior SOTA \small{(0.2\%)}&70.34&44.46&-&-&-&-\\
    \cellcolor{LightCyan}SNORD \small{(0.2\%)}&\cellcolor{LightCyan}\textbf{76.83}&\cellcolor{LightCyan}\textbf{49.74}&\cellcolor{LightCyan}-&\cellcolor{LightCyan}-&\cellcolor{LightCyan}-&\cellcolor{LightCyan}-\\
    \midrule
    Prior SOTA \small{(5\%)}&79.07&48.01&42.81&21.89&-&-\\
    \cellcolor{LightCyan}SNORD \small{(5\%)}&\cellcolor{LightCyan}\textbf{81.96}&\cellcolor{LightCyan}\textbf{49.29}&\cellcolor{LightCyan}\textbf{48.09}&\cellcolor{LightCyan}\textbf{23.42}&\cellcolor{LightCyan}-&\cellcolor{LightCyan}-\\
    \midrule
    Prior SOTA \small{(10\%)}&78.34&48.56&45.64&23.79&31.96&\ \ 9.26\\
    \cellcolor{LightCyan}SNORD \small{(10\%)}&\cellcolor{LightCyan}\textbf{82.73}&\cellcolor{LightCyan}\textbf{49.14}&\cellcolor{LightCyan}\textbf{52.03}&\cellcolor{LightCyan}\textbf{23.94}&\cellcolor{LightCyan}\textbf{41.70}&\cellcolor{LightCyan}\textbf{15.26}\\
    \bottomrule
    \end{tabular}
}
\caption{\textbf{More Comparisons.} We conduct a further comparison of SNORD against previous state-of-the-art SSL adversarial training methods and the fully adversarial supervision method, ADR \cite{wu2023annealing}, as well. Our findings demonstrate that SNORD not only outperforms all existing SSL adversarial training approaches across various settings but also significantly closes the gap in AutoAttack robustness (AA). 
}

\label{tab:sup_comp}
\end{table}

\subsection{Discussions}
\label{sec:abl}

\input{tables/abl_new}

\noindent \textbf{Ablation Studies.} We conducted comprehensive ablation studies to assess the efficacy of our proposed components on both the CIFAR-10 and CIFAR-100 datasets, as presented in \Cref{tab:Component_Ablation}. The results exhibit clear improvements when transitioning from standard training (row (a)) to advanced semi-supervised learning (SSL) algorithms (row (b)) under limited labeled data. The introduction of our NAR and ORD modules, individually illustrated in rows (c) and (d), yields evident enhancements in both performance and robustness over the baseline (row (b)). Notably, the combination of both NAR and ORD modules in row (e) showcases the most favorable outcomes in two prominent robustness evaluation benchmarks. However, a marginal performance drop in standard accuracy is observed in row (e) in comparison to applying NAR or ORD individually (rows (c) and (d)). This minor decline could be attributed to a cumulative effect of smoothing operations stemming from the combined deployment of NAR and ORD modules, aligning with insights from prior research \cite{zi2021revisiting}. Furthermore, we evaluate the efficacy of our NAR and ORD modules in conjunction with established RST-based methods like ACL and DynACL++. By intentionally excluding advanced SSL algorithms in favor of improved pseudo labels, our approach consistently yields superior results in \Cref{tab:diff_pretraining}, firmly underscoring the value of NAR and ORD.

\begin{figure}[!b]
    \centering
    \includegraphics[width=\linewidth]{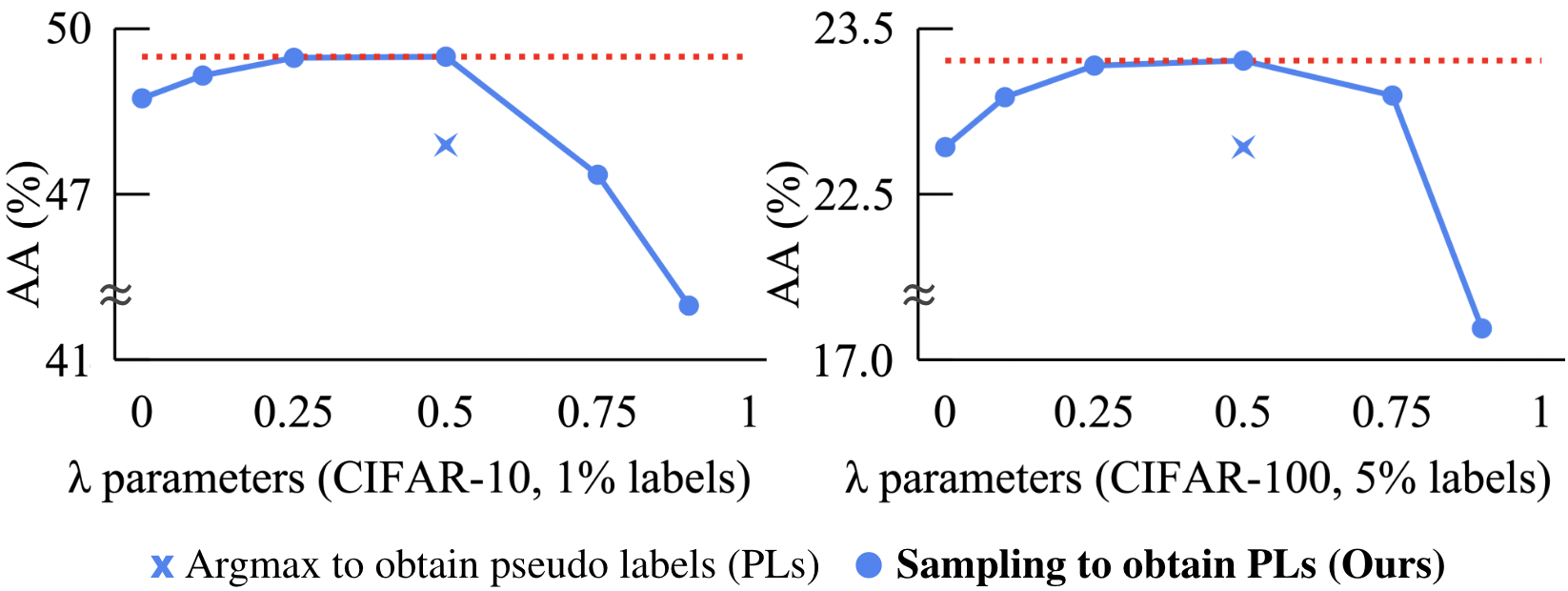}
    \caption{\textbf{Sensitivity analyses of our NAR methods.} The result underscores the importance of using sampling rather than argmax to obtain pseudo labels and the robustness of our approach over a wide range of hyper-parameters.
    }
    \label{fig:fig4}
\end{figure}

Additionally, we delve deeper into the mechanics of our proposed NAR method. Illustrated in \Cref{fig:fig4}, the hyper-parameter $\lambda$, responsible for harmonizing one-hot labels with predicted distributions, manifests consistent performance across a substantial range of values ($0.25 \sim 0.5$) on both datasets, affirming the robustness of our method. This observation further confirms the limitations inherent in the adversarial finetuning strategy employed by ACL and DynACL++, as the classifier's proficiency leans more toward one-hot labels than sole reliance on soft predicted distributions. Moreover, the comparison between our sampling process in equation (3), and the use of $\mathrm{argmax}(\cdot)$ function to obtain $y_{PL}$ (indicated by the X-mark in the figures) clearly demonstrates the superiority of our sampling approach. This underscores the importance of employing sampling to enhance label diversity and mitigate noise inherent in noisy pseudo labels.

\vspace{0.7em}

\noindent \textbf{Applicability to Adversarial Pretraining Methods.} We extend our evaluation to assess the broader applicability of our approach by integrating it with existing adversarial pretraining methods. In this context, we initialize the robust model with adversarially pretrained weights from DynACL++ and subsequently perform adversarial finetuning with the NAR and ORD modules from an SSL-trained pseudo label generator over 30 epochs. As demonstrated in \Cref{tab:pretraining}, when pseudo label quality is high, such as achieving over 85\% precision with 1\% CIFAR-10 labeled data, the application of adversarial pretraining on top of our SNORD method yields the best results, consistent with findings in \cite{luo2023rethinking}. However, when the initial pseudo label quality is lower, as indicated by over 40\% error with 5\% CIFAR-100 annotations, the effectiveness of initializing the network with robust pretrained weights is negligible compared to direct adversarial training from scratch using our SNORD framework. This highlights that in the realm of SSL adversarial training, the impact of robust pretrained models may be constrained when finetuning with noisy data. Instead, the absence of a robust pseudo label generator or an effective method to handle noisy training data can lead to a remarkable performance drop, as evident in the comparison between the first and second rows of the table.

\begin{table}[!t]
\centering
\small
\setlength\tabcolsep{3pt}
\resizebox{1.0\linewidth}{!}{
\begin{tabular}{c|ccc|ccc}
\toprule 
\multirow{2}{*}{\begin{tabular}[c]{@{}c@{}}Methods\end{tabular}} & \multicolumn{3}{c|}{CIFAR-10 1\%} & \multicolumn{3}{c}{CIFAR-100 5\%}\\
& \tabincell{c}{SA} & \tabincell{c}{RA} & \tabincell{c}{AA} & \tabincell{c}{SA} & \tabincell{c}{RA} & \tabincell{c}{AA}\\
\midrule
ACL&75.45&50.59&46.18&42.57&25.64&\textbf{21.90}\\
+NAR+ORD&\textbf{81.21}&\textbf{52.50}&\textbf{48.16}&\textbf{50.67}&\textbf{26.22}&21.83\\
\midrule
DynACL++&76.77&51.30&46.95&42.81&25.93&21.89\\
+NAR+ORD&\textbf{80.54}&\textbf{52.30}&\textbf{48.45}&\textbf{48.92}&\textbf{26.58}&\textbf{22.20}\\

\bottomrule  
\end{tabular}%
}

\caption {\textbf{Effectiveness of NAR and ORD.}  Without the use of SSL-trained pseudo label generator, we showcase the benefit of NAR and ORD on top of prior RST-based pipelines.}
\label{tab:diff_pretraining}%
\end{table}%

\begin{table}[!t]
\centering
\small
\setlength\tabcolsep{3pt}
\resizebox{1.0\linewidth}{!}
{
\begin{tabular}{c|ccc|ccc}
\toprule 
\multirow{2}{*}{\begin{tabular}[c]{@{}c@{}}Methods\end{tabular}} & \multicolumn{3}{c|}{CIFAR-10 1\%} & \multicolumn{3}{c}{CIFAR-100 5\%}\\
& \tabincell{c}{SA} & \tabincell{c}{RA} & \tabincell{c}{AA} & \tabincell{c}{SA} & \tabincell{c}{RA} & \tabincell{c}{AA}\\
\midrule
DynACL++&76.77&51.30&46.95&42.81&25.93&21.89\\
SNORD&80.60&53.22&49.47&\textbf{48.09}&27.05&\textbf{23.42}\\
\midrule
DynACL++ \& SNORD&\textbf{82.17}&\textbf{53.78}&\textbf{50.37}&48.00&\textbf{27.29}&23.19\\

\bottomrule  
\end{tabular}%
}

\caption {\textbf{Combination of SNORD with adversarial pretraining methods.} With high-quality pseudo labels on CIFAR-10, the combined utilization of a robust pretrained model and our SNORD mechanism yields enhanced outcomes in comparison to employing either one in isolation. Nonetheless, when confronted with severe noisy training data on CIFAR-100, the potential effectiveness of the pretrained model becomes constrained, underscoring the substantial significance of SNORD.}
\label{tab:pretraining}%
\end{table}%

%% file: tables/big_tab_C10.tex
\begin{table*}[t]
\centering
\setlength\tabcolsep{2pt}
\resizebox{1.0\linewidth}{!}
{
\begin{tabular}{c|ccc|ccc|ccc|ccc|ccc|ccc}
\toprule
% \multicolumn{19}{c}{CIFAR-10 Test Accuracy (\%) under SSL settings}\\
% \midrule
\multicolumn{1}{c|}{
\multirow{2}{*}{\begin{tabular}[c]{@{}c@{}}Methods\end{tabular}}}&\multicolumn{3}{c|}{0.1\% labels}&\multicolumn{3}{c|}{0.2\% labels}&\multicolumn{3}{c|}{0.5\% labels}&\multicolumn{3}{c|}{1\% labels}
&\multicolumn{3}{c|}{5\% labels}
&\multicolumn{3}{c}{10\% labels}\\
&SA&RA&AA&SA&RA&AA&SA&RA&AA&SA&RA&AA&SA&RA&AA&SA&RA&AA\\

\midrule
% TRADES~\cite{zhang2019theoretically}&15.05$\pm$2.75&11.88$\pm$1.68&9.76$\pm$4.65&17.44$\pm$1.11&11.91$\pm$1.44&11.57$\pm$1.50&61.59$\pm$1.09&32.70$\pm$0.44&29.01$\pm$0.43\\
% TRADES &-&-&-&-&-&-&-&-&-&-&-&-&-&-&-\\
UAT++ &24.87&13.28&11.25&31.79&19.88&17.93&41.47&25.19&22.29&49.75&30.74&27.16&70.41&45.83&42.69&76.34&49.57&46.11\\
% RST &&&&-&-&-&-&-&-&-&-&-&-&-&-&-&-&-\\
% UAT++ \cite{alayrac2019labels}&-&-&-&-&-&-&-&-&-\\
% UAT++ \cite{alayrac2019labels}&18.34$\pm$2.25&14.75$\pm$0.70&14.49$\pm$0.61&-&-&-&56.88$\pm$1.93&37.51$\pm$1.23&35.07$\pm$1.11\\
% SRT \cite{li2022semi}&20.49$\pm$1.40&9.43$\pm$2.10&8.65$\pm$2.01&34.93$\pm$0.28&17.26$\pm$0.68&16.13$\pm$0.67&62.75$\pm$0.64&33.47$\pm$0.23&31.89$\pm$0.26\\
% ACL &48.25&35.08&31.00&68.38&46.43&41.89&71.99&50.20&45.26&75.45&50.59&46.18&77.77&51.21&47.11&76.37&51.73&47.14\\
ACL &48.02&35.03&31.04&68.38&46.43&41.89&71.99&50.20&45.26&75.45&50.59&46.18&77.77&51.21&47.11&76.37&51.73&47.14\\
% DynACL++ (GG)&70.34&50.14&44.46&69.92&51.03&45.49&70.58&52.36&46.85(C5)&79.07&51.35&48.01&78.84&52.08&48.43\\
DynACL++&64.34&47.31&44.27&70.34&50.14&44.46&69.92&51.03&45.49&76.77&51.30&46.95&79.07&51.35&48.01&78.34&\textbf{53.00}&48.56\\
% \cellcolor{LightCyan}SNORD (Ours)&\cellcolor{LightCyan}73.20&\cellcolor{LightCyan}50.94&\cellcolor{LightCyan}48.12&\cellcolor{LightCyan}77.80&\cellcolor{LightCyan}52.29&\cellcolor{LightCyan}48.97&\cellcolor{LightCyan}79.59&\cellcolor{LightCyan}53.03&\cellcolor{LightCyan}49.04&\cellcolor{LightCyan}80.28&\cellcolor{LightCyan}53.27&\cellcolor{LightCyan}49.26&\cellcolor{LightCyan}81.50&\cellcolor{LightCyan}52.86&\cellcolor{LightCyan}48.74&\cellcolor{LightCyan}82.59&\cellcolor{LightCyan}53.07&\cellcolor{LightCyan}49.47\\
\cellcolor{LightCyan}\textbf{SNORD (Ours)}&\cellcolor{LightCyan}\textbf{71.71}&\cellcolor{LightCyan}\textbf{50.07}&\cellcolor{LightCyan}\textbf{47.12}&\cellcolor{LightCyan}\textbf{76.83}&\cellcolor{LightCyan}\textbf{53.28}&\cellcolor{LightCyan}\textbf{49.74}&\cellcolor{LightCyan}\textbf{80.99}&\cellcolor{LightCyan}\textbf{53.46}&\cellcolor{LightCyan}\textbf{49.48}&\cellcolor{LightCyan}\textbf{80.60}&\cellcolor{LightCyan}\textbf{53.22}&\cellcolor{LightCyan}\textbf{49.47}&\cellcolor{LightCyan}\textbf{81.96}&\cellcolor{LightCyan}\textbf{52.90}&\cellcolor{LightCyan}\textbf{49.29}&\cellcolor{LightCyan}\textbf{82.73}&\cellcolor{LightCyan}52.97&\cellcolor{LightCyan}\textbf{49.14}\\

% \midrule
% Oracle Sup-AT&-&-&-&-&-&-&-&-&-&-&-&-&-&-&-\\
% 100\% labels AT&80.91$\pm$0.49&52.30$\pm$0.07&47.95$\pm$0.24&80.91$\pm$0.49&52.30$\pm$0.07&47.95$\pm$0.24&80.91$\pm$0.49&52.30$\pm$0.07&47.95$\pm$0.24\\
\bottomrule
\end{tabular}
}
    % \vspace{-0.3em}
    \caption{\textbf{CIFAR-10 test accuracy (\%) under SSL settings.} SA, RA, and AA denote standard accuracy, PGD-20 robust accuracy, and AutoAttack robust accuracy, respectively. For baselines, we report the maximum value of their official number and our reproduced results.}
    
    % Comparison between various active learning methods on the HAM-10000 dataset. The \red{red} and \blue{blue} color indicate the best and the second best result respectively.}
    \label{tab:main_result}
    % \vspace{-0.7em}
\end{table*}

%% file: tables/big_tab_C100.tex
\begin{table*}[t]
\centering
\setlength\tabcolsep{2pt}
\resizebox{0.75\linewidth}{!}
{
\begin{tabular}{c|ccc|ccc|ccc|ccc}
\toprule
% \multicolumn{13}{c}{CIFAR-100 Test Accuracy (\%) under SSL settings}\\
% \midrule
\multicolumn{1}{c|}{
\multirow{2}{*}{\begin{tabular}[c]{@{}c@{}}Methods\end{tabular}}}&\multicolumn{3}{c|}{1\% labels}&\multicolumn{3}{c|}{2\% labels}&\multicolumn{3}{c|}{5\% labels}&\multicolumn{3}{c}{10\% labels}\\
&SA&RA&AA&SA&RA&AA&SA&RA&AA&SA&RA&AA\\

\midrule
% TRADES~\cite{zhang2019theoretically}&15.05$\pm$2.75&11.88$\pm$1.68&9.76$\pm$4.65&17.44$\pm$1.11&11.91$\pm$1.44&11.57$\pm$1.50&61.59$\pm$1.09&32.70$\pm$0.44&29.01$\pm$0.43\\
% TRADES &-&-&-&-&-&-&-&-\\
UAT++ &10.43&4.14&3.59&17.48&6.52&5.66&25.12&12.13&10.41&38.63&18.86&16.01\\
% RST &&&&-&-&-&-&-&-&-&-\\
ACL &27.87&19.39&16.51&32.10&19.83&17.70&42.57&25.64&21.90&44.05&26.78&22.33\\
DynACL++ &35.34&21.55&18.43&32.92&21.33&18.48&42.81&25.93&21.89&45.64&27.98&23.79\\
% \cellcolor{LightCyan}SNORD (Ours) &\cellcolor{LightCyan}34.64&\cellcolor{LightCyan}22.59(CY)&\cellcolor{LightCyan}19.52&\cellcolor{LightCyan}43.88&\cellcolor{LightCyan}26.32(CY)&\cellcolor{LightCyan}22.22&\cellcolor{LightCyan}50.91&\cellcolor{LightCyan}28.96(C5)&\cellcolor{LightCyan}24.33\\
% \cellcolor{LightCyan}SNORD (Ours) &\cellcolor{LightCyan}36.36&\cellcolor{LightCyan}21.65&\cellcolor{LightCyan}18.98&\cellcolor{LightCyan}43.86&\cellcolor{LightCyan}26.06&\cellcolor{LightCyan}22.25&\cellcolor{LightCyan}48.02&\cellcolor{LightCyan}26.90&\cellcolor{LightCyan}22.70&\cellcolor{LightCyan}51.70&\cellcolor{LightCyan}28.74&\cellcolor{LightCyan}24.07\\
\cellcolor{LightCyan}\textbf{SNORD (Ours)}&\cellcolor{LightCyan}\textbf{35.44}&\cellcolor{LightCyan}\textbf{22.08}&\cellcolor{LightCyan}\textbf{19.00}&\cellcolor{LightCyan}\textbf{44.14}&\cellcolor{LightCyan}\textbf{26.14}&\cellcolor{LightCyan}\textbf{22.46}&\cellcolor{LightCyan}\textbf{48.09}&\cellcolor{LightCyan}\textbf{27.05}&\cellcolor{LightCyan}\textbf{23.42}&\cellcolor{LightCyan}\textbf{52.03}&\cellcolor{LightCyan}\textbf{28.27}&\cellcolor{LightCyan}\textbf{23.94}\\
% \midrule
% Oracle Sup-AT&00.00&00.00&00.00&00.00&00.00&00.00&00.00&00.00&00.00\\
\bottomrule
\end{tabular}
}
    \caption{\textbf{CIFAR-100 test accuracy (\%) under SSL settings.} Our method showcased superior performance over all existing baseline methods across diverse labeling budgets, establishing its dominance across all SSL scenarios.}
    
    % Comparison between various active learning methods on the HAM-10000 dataset. The \red{red} and \blue{blue} color indicate the best and the second best result respectively.}
    \label{tab:c100_result}
    % \vspace{-0.7em}
\end{table*}

%% file: tables/big_tab_TIN.tex
\begin{table}[h!]
\centering
\setlength\tabcolsep{2pt}
\resizebox{\linewidth}{!}
{
\begin{tabular}{c|ccc|ccc}
\toprule
% \multicolumn{7}{c}{TingImageNet-200 Test Accuracy (\%) under SSL settings}\\
% \midrule
\multicolumn{1}{c|}{
\multirow{2}{*}{\begin{tabular}[c]{@{}c@{}}Methods\end{tabular}}}&\multicolumn{3}{c|}{10\% labels}&\multicolumn{3}{c}{20\% labels}\\
&SA&RA&AA&SA&RA&AA\\

\midrule
% TRADES~\cite{zhang2019theoretically}&15.05$\pm$2.75&11.88$\pm$1.68&9.76$\pm$4.65&17.44$\pm$1.11&11.91$\pm$1.44&11.57$\pm$1.50&61.59$\pm$1.09&32.70$\pm$0.44&29.01$\pm$0.43\\
% TRADES &-&-&-&-&-&-&-&-\\
UAT++ &31.96&12.98&9.26&33.36&13.30&9.88\\
% RST &&&&-&-&-&-&-&-&-&-\\
% ACL &27.87&19.39&16.51&32.10&cml9&17.70\\
% DynACL++ &35.34&-&18.43&32.92&cml9&18.48\\
% \cellcolor{LightCyan}SNORD (Ours) &\cellcolor{LightCyan}34.64&\cellcolor{LightCyan}22.59(CY)&\cellcolor{LightCyan}19.52&\cellcolor{LightCyan}43.88&\cellcolor{LightCyan}26.32(CY)&\cellcolor{LightCyan}22.22&\cellcolor{LightCyan}50.91&\cellcolor{LightCyan}28.96(C5)&\cellcolor{LightCyan}24.33\\
% \cellcolor{LightCyan}SNORD (Ours) &\cellcolor{LightCyan}36.36&\cellcolor{LightCyan}21.65&\cellcolor{LightCyan}18.98&\cellcolor{LightCyan}43.86&\cellcolor{LightCyan}26.06&\cellcolor{LightCyan}22.25&\cellcolor{LightCyan}48.02&\cellcolor{LightCyan}26.90&\cellcolor{LightCyan}22.70&\cellcolor{LightCyan}51.70&\cellcolor{LightCyan}28.74&\cellcolor{LightCyan}24.07\\
\cellcolor{LightCyan}\textbf{SNORD (Ours)}&\cellcolor{LightCyan}\textbf{41.70}&\cellcolor{LightCyan}\textbf{20.02}&\cellcolor{LightCyan}\textbf{15.26}&\cellcolor{LightCyan}\textbf{46.84}&\cellcolor{LightCyan}\textbf{22.00}&\cellcolor{LightCyan}\textbf{16.70}\\
% \midrule
% Oracle Sup-AT&00.00&00.00&00.00&00.00&00.00&00.00&00.00&00.00&00.00\\
\bottomrule
\end{tabular}
}
    \caption{\textbf{TinyImageNet-200 test accuracy (\%) under SSL settings.} We only compare our method to the basic RST method (UAT++), given that previous adversarial pretraining approaches had refrained from such a large-scale dataset.}
    
    % Comparison between various active learning methods on the HAM-10000 dataset. The \red{red} and \blue{blue} color indicate the best and the second best result respectively.}
    \label{tab:tin_result}
    % \vspace{-0.7em}
\end{table}

%% file: tables/abl_new.tex
\begin{table}[!b]

\centering
\small
\setlength\tabcolsep{2pt}\resizebox{1.0\linewidth}{!}{
\begin{tabular}{c|ccc|ccc|ccc}
\toprule 
\multicolumn{1}{c|}{} & \multicolumn{3}{c|}{Components} & \multicolumn{3}{c|}{CIFAR-10 1\%} & \multicolumn{3}{c}{CIFAR-100 5\%}\\
\midrule
& \tabincell{c}{\footnotesize{SSL algo.}} & \tabincell{c}{\footnotesize{NAR}} & \tabincell{c}{\footnotesize{ORD}} & \tabincell{c}{SA} & \tabincell{c}{RA} & \tabincell{c}{AA} & \tabincell{c}{SA} & \tabincell{c}{RA} & \tabincell{c}{AA}\\
\midrule
(a) &  & &  & 55.58 & 34.52 & 31.68 & 33.90 & 17.19 & 14.47 \\
(b) & \checkmark & &  & 79.91 & 52.06 & 48.72 & 48.57 & 26.44 & 22.78 \\
(c) & \checkmark & \checkmark &  & \textbf{81.81} & 52.50 & 49.09  & \textbf{49.84} & 26.92 & 23.13 \\
% (d) & \checkmark & & \checkmark & 81.55 & 52.07 & 48.34 & \textbf{49.88} & 27.16 & 23.50 \\
(d) & \checkmark & & \checkmark & 81.55 & 52.07 & 48.34 & 48.52 & 26.85 & 23.34 \\
% (e) &  & \checkmark & \checkmark &00.00&00.00&00.00&00.00&00.00&00.00\\
% (f) & \checkmark & \checkmark & \checkmark & 80.16 & 52.82 & 48.88\\
(e) & \checkmark & \checkmark & \checkmark & 80.60 & \textbf{53.22} & \textbf{49.47} & 48.29 & \textbf{27.05} & \textbf{23.42} \\
% (b) &  & \checkmark &  &  62.75 & 68.81 & 69.82\\
% (c) &  &\checkmark & \checkmark  & \textbf{64.03} & 69.38 & 70.69\\
% %\cline{1-1} \cline{3-22}    
% (d) & \checkmark&\checkmark &  &63.30 & 69.66 & 71.15\\
% %\cline{1-1} \cline{3-22}    
% (e) & \checkmark&\checkmark & \checkmark  & \textbf{64.03} &  \textbf{69.86} & \textbf{71.25} \\

\bottomrule  
\end{tabular}%
}

\caption {\textbf{Ablation studies.} By harnessing the capabilities of our developed NAR and ORD modules, in conjunction with an SSL-trained pseudo label generator, we are able to achieve the optimal results for both RA and AA.}
% \vspace{-0.7em}
\label{tab:Component_Ablation}%
\end{table}%

%% file: sections/5_conclusion.tex
We present SNORD, a simple, effective, and general SSL adversarial training framework in the semi-supervised learning paradigm. Instead of developing new adversarial pertaining algorithms as a lot of prior work, we revised the widely-used RST-based methods and pointed out that the bottleneck to this problem is the quality of pseudo-labels and the management of noisy training data. With the aid of existing and our developed SSL techniques, SNORD demonstrates a substantial performance advantage over conventional RST-based approaches across multiple well-established benchmarks, regardless of the usage of adversarial pretrained models. This success paves the way for a novel approach to label-efficient and adversarial robust visual recognition systems.

%% file: sections/X_supp_arXiv.tex
\clearpage
\maketitlesupplementary
\appendix

\section{Implementation Details}
\label{sec:impl}

In this section, we present the specifics of our implementation, covering the computing infrastructure used and the training process details. We intend to release our source code once the paper is accepted.

\subsection{Computing Infrastructure}

All experiments were conducted on a personal computer with an 8-core CPU and an NVIDIA RTX3090 GPU. The operating system was Ubuntu 20.04, and we implemented the system using Python 3.7. For efficient computations, we leveraged CUDA 11.3 and employed PyTorch 1.11.0 as our deep learning framework.

\subsection{Training Details}

Our SNORD approach involves training a pseudo label generator through semi-supervised learning (SSL) and a robust model adversarial trained by our noise-aware manner. We trained on CIFAR-10/100 images of size $32 \times 32$ and on TinyImageNet-200 images of size $64 \times 64$. The data generation method is detailed in Section 4.1 of the main manuscript, where we adopt the data split from \cite{rice2020overfitting} for CIFAR-10/100 and rely on the official split for TinyImageNet-200. We constructed class-balanced SSL datasets following established methods such as \cite{sohn2020fixmatch, ACL2020, luo2023rethinking}.

\vspace{0.7em}

\noindent \textbf{Training Pseudo Label Generator.} In our experiments, the SSL-trained model was trained using the FixMatch algorithm \cite{sohn2020fixmatch} with a ResNet-18 \cite{he2016deep} backbone. The hyperparameters for training our pseudo label generator are detailed in \Cref{tab:hyper}. In this table, $\tau$ represents the confidence threshold, and $\mu$ is the ratio of unlabeled data to labeled data. The training incorporated the SGD optimizer with an initial learning rate of 0.03, momentum of 0.9, and Nesterov momentum enabled, consistent for the three datasets. The weight decay for CIFAR-10, CIFAR-100, and TinyImageNet-200 were set to 1e-3, 2e-4, and 1e-4, respectively. We employed the original cosine annealing scheduler for adaptive learning rate adjustment. Other parameters and data augmentation methods remained consistent with those in the original paper.

\begin{table}[h]
\resizebox{1.0\linewidth}{!}{
\centering

\begin{tabular}{c|c|c|c}
\toprule
             & CIFAR-10 & CIFAR-100 & TinyImageNet-200 \\
\midrule
$\tau$          & \multicolumn{3}{c}{0.95}                \\
$\mu$           & \multicolumn{3}{c}{5}                   \\
$B$            & \multicolumn{3}{c}{64}                  \\
$lr$           & \multicolumn{3}{c}{0.03}                \\
\midrule
weight decay & 0.001    & 0.0002    & 0.0001 \\
\bottomrule

\end{tabular}

}
    \caption{Hyperparameters for training the pseudo label generator on the three datasets.}
    \label{tab:hyper}
\end{table}

\vspace{0.7em}

\noindent \textbf{Training Robust Model.} After detailing the pseudo label generator's training, we proceed to explain the adversarial training of the robust model. The robust model's training commenced at the 128th and 256th epochs of the pseudo label generator on CIFAR and TinyImageNet datasets, respectively, as the standard model's performance was satisfactory.

The robust model was trained for 200 epochs for the CIFAR-10/100 and 80 epochs for the TinyImageNet-200 following existing work \cite{pang2021bag,wu2023annealing}. We optimize the model with the SGD optimizer, initialized with a learning rate of 0.1 and a weight decay of 5e-4, in line with prior studies \cite{pang2021bag}. The learning rate was managed through a piecewise learning rate scheduler, with a reduction factor of 0.1 at the 50\% and 75\% epochs of the total robust model training duration \cite{rice2020overfitting}. Data augmentation included basic techniques such as random cropping with padding and random horizontal flipping, consistent with prior studies \cite{pang2021bag, rice2020overfitting, ACL2020}.

Notably, we set the noise-aware label rectification parameter $\lambda$ in equation (2) of the main paper to 0.5. The parameter's sensitivity was evaluated in Figure 4 of the main manuscript. We set the $\beta$ parameter in equation (5) of the main paper to $6$, following TRADES \cite{zhang2019theoretically}. Additionally, we applied the proposed ORD method from the standard model to the robust model every 5 epochs, indicating that we update the robust model once after 5 standard model updates. This strategy was chosen as we observed that allowing the standard model to update for multiple epochs enhanced label diversity and overall performance.

\vspace{0.7em}

\noindent \textbf{Ensuring Fair Comparison.} To ensure consistency, each experimental setting, including all the baseline methods, was executed once with the same random seed, as the observed variation on two major robust evaluation metrics was minimal as shown in \Cref{tab:run}. Our reported results, following well-established protocols \cite{rice2020overfitting}, are derived from the best checkpoint, selected based on the highest Robustness Accuracy (RA) on the validation set. 

\begin{table}[!h]
\centering

\begin{tabular}{c|ccc|ccc}
\toprule 
\multirow{2}{*}{\begin{tabular}[c]{@{}c@{}}Methods\end{tabular}} & \multicolumn{3}{c|}{CIFAR-10 5\%} & \multicolumn{3}{c}{CIFAR-100 5\%}\\
& \tabincell{c}{SA} & \tabincell{c}{RA} & \tabincell{c}{AA} & \tabincell{c}{SA} & \tabincell{c}{RA} & \tabincell{c}{AA}\\
\midrule
Run 1&81.66&52.91&49.25&48.29&27.25&23.30\\
Run 2&81.27&52.90&49.29&48.09&27.05&23.42\\

\bottomrule  
\end{tabular}%

\caption {Performance variation with different random seeds.}
\label{tab:run}
\end{table}%

Note that to accelerate the whole training process, we harnessed automatic mixed precision (AMP) training. Additionally, we implemented early stopping as deemed necessary within our approach. Conversely, for other methods, we utilized their original implementations with 32-bit full precision training. This approach was implemented to efficiently manage computational resources while ensuring fair comparison across methods. 

The overall training time of our SNORD method, including both the pseudo label generator and the robust model, for CIFAR-10/100 amounted to approximately 32 hours. This timeframe is quite comparable to the requirements of our baseline methods ACL \cite{ACL2020} and DynACL++ \cite{luo2023rethinking}, which involve around 30 hours for adversarial pretraining and 2 to 3 hours for adversarial fine-tuning. As for the TinyImageNet-200 dataset, due to its greater complexity, the training time of SNORD was extended to approximately 3.5 days to effectively train the model.